\documentclass[conference]{IEEEtran}
\usepackage[numbers]{natbib}
\input epsf
\usepackage{graphicx}
\usepackage{import}
\usepackage{lipsum}
\usepackage[]{hyperref}
\hypersetup{
  colorlinks   = true, 
  urlcolor     = black, 
  linkcolor    = black, 
  citecolor    = black}
\usepackage[ruled]{algorithm2e}
\LinesNumbered
\SetAlgoSkip{bigskip}
\usepackage{natbib}
\usepackage[caption=false, font=footnotesize]{subfig}
\usepackage{amsmath}
\usepackage{booktabs}

\usepackage{xcolor}
\usepackage{soul}

\begin{document}

\title{\LARGE Action Capsules: Human Skeleton Action Recognition}
\author{\authorblockN{Ali Farajzadeh Bavil, Hamed Damirchi, Hamid D. Taghirad, \em{Senior Member, IEEE}\\
Advanced Robotics and Automated Systems (ARAS),\\
Faculty of Electrical Engineering, \\
K. N. Toosi University of Technology, Tehran, lran.\\}
Email: alifarajzadeh@email, hdamirchi@email, taghirad@kntu.ac.ir}

\maketitle

\begin{abstract}
Due to the compact and rich high-level representations offered, skeleton-based human action recognition has recently become a highly active research topic. Previous studies have demonstrated that investigating joint relationships in spatial and temporal dimensions provides effective information critical to action recognition. However, effectively encoding global dependencies of joints during spatio-temporal feature extraction is still challenging. In this paper, we introduce \emph{Action Capsule} which identifies action-related key joints by considering the latent correlation of joints in a skeleton sequence. We show that, during inference, our end-to-end network pays attention to a set of joints specific to each action, whose encoded spatio-temporal features are aggregated to recognize the action. Additionally, the use of multiple stages of action capsules enhances the ability of the network to classify similar actions.
Consequently, our network outperforms the state-of-the-art approaches on the N-UCLA dataset and obtains competitive results on the NTURGBD dataset. This is while our approach has significantly lower computational requirements based on GFLOPs measurements.

\end{abstract}


\begin{figure*}[!ht]
    \centering
    \includegraphics[scale=0.8]{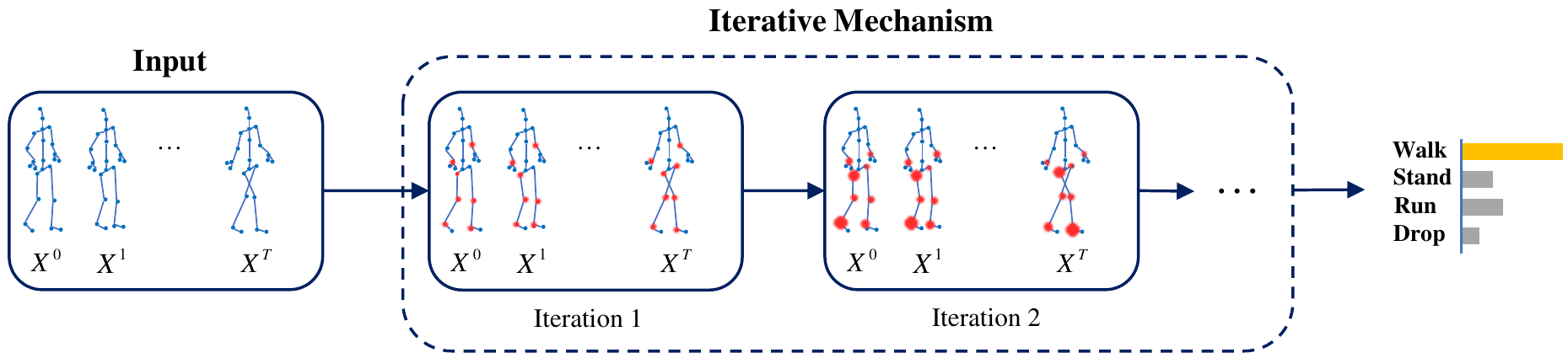}
    \caption{Human skeleton action recognition using the Action Capsules. The method uses an iterative mechanism to select contextual joints of a skeleton sequence. Through iterations, the significance of joints is refined to achieve a set of discriminative action-related joints. Finally encoded spatio-temporal features of selected joints are aggregated to classify actions.}
\end{figure*}

\section{Introduction}
Action recognition is considered one of the prominent problems in computer vision due to its applications in human activity understanding~\cite{simonyan2014two}, intelligent surveillance systems~\cite{gaur2011string}, human-machine interaction~\cite{gui2018teaching}, and human social behavior analysis~\cite{bagautdinov2017social}. Each application requires the use of a particular modality or the combination thereof, such as RGB images~\cite{feichtenhofer2019slowfast}, sensor data~\cite{damirchi2020arc}, depth maps~\cite{xu2017lie}, optical flow~\cite{wang2016temporal}, and body skeletons~\cite{shahroudy2016ntu}, to name a few. Among these widely used modalities, skeleton-based action recognition, where sequences of 2D or 3D coordinated human joint locations are used as input data, has increasingly attracted researchers since it provides compact data with rich and high-level features for action recognition while being robust to intra-class variations as well as elaborated backgrounds of videos~\cite{li2018co}.

Recently, several studies~\cite{yan2018spatial, plizzari2021skeleton, shi2020decoupled} have shown that exploring the spatial and temporal features of the skeleton sequence is of paramount importance for action recognition. Deep learning-based architectures such as the Recurrent Neural Network (RNN), Convolutional Neural Network (CNN), and Graph Convolutional Network (GCN) are among the widely used structures that attempted to accomplish this goal by transforming raw skeleton data into a vector sequence, pseudo-image, and graph, respectively.  
However, they leave out the exploration of the global spatial and temporal relationship of joints. This is especially detrimental for some activities that require the mutual consideration of high-level features extracted from multiple joints.

Although CNNs perform well in terms of extracting temporal features~\cite{lea2017temporal}, they fail to retain the spatial relationships~\cite{wang2018non} among joints involved during an action owing to their structural deficiencies. These include discarding spatial information by using pooling layers and generating only local features at each layer, which leads to information loss concerning the relationship between the joints.
In this regard, it is reasonable to model actions not only by temporal features of joints but also by their relative location and relationship.

CapsNet~\cite{sabour2017dynamic} is an emerging type of neural network that proves useful in addressing some flaws associated with traditional convolutional neural networks. CapsNet defines a type of neuron called the capsule. Capsules operate on vectors and characterize patterns in input, which defines the likelihood of its presence as well as encodes its relation to other capsules. 
Accordingly, activation of a capsule in the lower levels of the network can be interpreted as the presence of a specific movement of the joint that a sequence of data belongs. Meanwhile, the activation of a higher-level capsule can correspond to the presence of an overall action once all the joints are considered mutually.

By leveraging the above observations, in this paper, we propose a new model called \emph{Action Capsule Network} for skeleton-based action recognition. 
As basic components, we propose \emph{Residual Temporal Convolution Neural Network (Res-TCN)} to encode temporal features associated with each joint hierarchically, and \emph{CapsNet} to dynamically focus on a set of critical joints. The main aspects of our contributions are as follows:
\begin{itemize}
\item Human actions are diverse within each action class, since subjects may perform the same action differently. We propose a scheme of extracting spatial features based on output features of hierarchical temporal convolutions, which results in temporal shift-invariance.
\item We propose a novel structure for skeleton-based action recognition based on capsule neural networks, which classifies actions by learning to dynamically attend to features of pivotal joints, specific for each action.
\item Human actions are occasionally reciprocal. Our method represents the relationship between joints without taking into account who the joints belong to, which to the best of our knowledge has not been attempted before.
\item Our method outperforms state-of-the-art methods in skeleton-based action recognition approaches while requiring a less computational cost.
\end{itemize}
\section{Related works}
From a categorical perspective, previous works include methods that rely on handcrafted features or deep learning-based methods. Using handcrafted features, as an early approach to skeleton-based action recognition, typically involves embedding the physical relationships between body parts. ~\cite{vemulapalli2014human}, represents rotations and translations of body parts in 3D space as SE(3) transformation matrices and then uses an SVM classifier to recognize the actions. Features offered by this category of methods suffer from disability in learning complex semantics, thereby resulting in poor performance. On the other hand, deep learning-based approaches have the potential to automatically extract both low-level and high-level features, making them highly desirable. The literature on this approach may be split into three categories based on the neural structure of the approach, namely, RNN-, CNN-, and GCN-based methods.

The RNN-based approaches regard skeleton data as a temporal sequence that is fed to an RNN. ~\cite{du2015hierarchical}, divided the human skeleton joints into five sets including the torso, arms, and legs which are then fed into five RNN sub-networks that are hierarchically fused in layers.
~\cite{zhang2017view} addressed the viewpoint variation problem in skeleton-based action recognition by proposing a view adaption module that automatically adjusts the observation viewpoint by transforming the skeleton into a new coordinate system.
~\cite{song2017end} devised a spatial-temporal attention mechanism for skeleton based action recognition. This mechanism operates by varying the degree of importance of each joint in a single frame while also weighting the information considered from different frames.
~\cite{liu2017global} introduced an explicit attention mechanism using a global context memory which is refined recurrently.

CNNs capture spatial information but need the input data to be put in a 2D array or 3D array structure similar to images. Therefore,~\cite{li2017skeleton, choutas2018potion, ludl2019simple} transformed the skeleton sequence to a pseudo image which can be classified by a CNN classifier.~\cite{li2017skeleton}, proposed an image mapping method independent of the translation and scale of the skeleton data.
These approaches, however, only consider adjacent joints when extracting features, and interpreting the co-occurrence of all joints is not possible in late layers. To this end, a framework for learning global co-occurrence features, developed by ~\cite{li2018co}, applied CNN to extract joint-level features individually, then treated each joint as a channel of features in a hierarchy of convolutional layers to learn co-occurrence features from all joints.

GCNs have drawn enormous attention in action recognition due to their ability in dealing with non-Euclidean data, including the skeletal data modeled by a graph in which joints are vertices and physical connections of the human body are edges. A static method called ST-GCN proposed by Yan et al.~\cite{yan2018spatial} was the first to recognize a human skeleton using GCN whose topology was fixed and predetermined based on the physical connections of the human body. GCN-based action recognition was made more flexible through dynamic methods~\cite{shi2020skeleton, li2019actional}, which dynamically adjusted the topology of GCN during inference.~\cite{li2019actional} captures richer dependencies between joints by introducing a generalized skeleton graph that uses an actional link inference module to capture action-related joint dependency and structural links to represent higher-order relationships between joints.

The aforementioned structure-aware approaches aggregate spatial features from local neighbors. In this case, however, the global relationship between joints is not explored, and there is no guarantee that the model will be able to explicitly learn the latent correlation between action-related joints that might be far from each other in the skeleton body, such as hands and feet that move together when walking.
We propose \emph{Activity-CapsNet}, where the structure allows for the relation between any sets of joints to be considered regardless of the proximity of the joints in the skeleton model.
\section{The Proposed Approach} \label{approach}
The general overview of our architecture is shown in ~\autoref{overall_architecture}.
In this section, the detailed architecture of \emph{Action Capsules Network} is introduced which uses \emph{Res-TCN} and \emph{CapsNet} to extract spatial-temporal features. The Res-TCN module, described in ~\autoref{TCN}, encodes data on the temporal dimension for each joint. Extracted features are then passed to CapsNet, described in ~\autoref{caps}, to encode the spatial relationship between joints and explore the co-occurrence of encoded temporal features. The overall architecture of the Activity-CapsNet network is described in ~\autoref{act-caps}.

\subsection{Residual Temporal Convolutional (Res-TCN) Block}
\label{TCN}
\begin{figure}[!ht]
    \centering
    \includegraphics[width=0.85\columnwidth,scale=1]{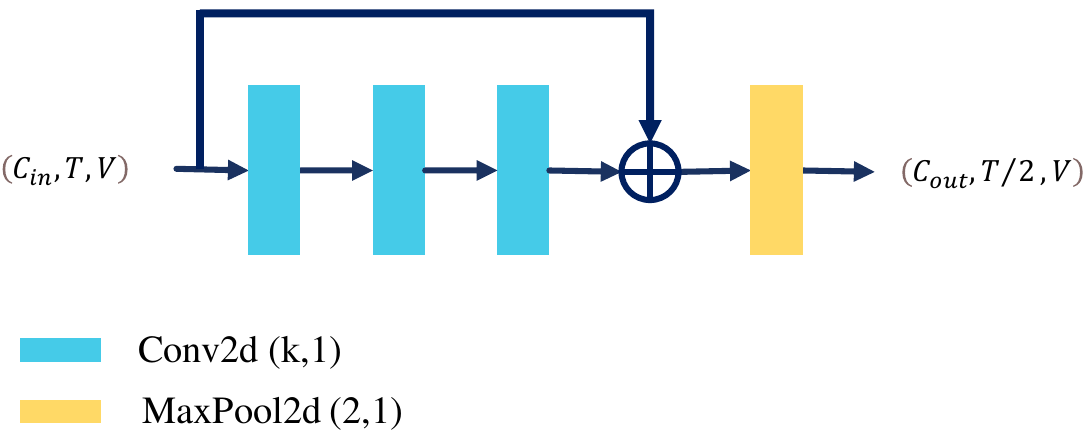}
    \caption{Structure of a Res-TCN layer in which $C, T, V$ denotes the number of channels, frames, and joints, respectively. Convolutional and MaxPlooling layers operate on the temporal dimension with $ k \times 1$ and $2 \times 1$ kernel sizes, respectively, with channels as the first dimension of input features. }
    \label{res-tcn}
\end{figure}

 A residual temporal convolution block as shown in ~\autoref{res-tcn}, consists of three consecutive temporal convolutions.
To perform convolution among the temporal dimension, a 2D convolution with a kernel size of $k \times 1$ has been conducted on a 
 feature map from the previous layer. We supplement the block with a residual connection to enhance the stability of the training and facilitate gradient propagation.
 
A max-pooling layer is placed after every Res-TCN to reduce temporal dimensionality, and mitigate the issues arising from differences in the chronology of joint movements across individuals for the same action.
Thus, temporal shift invariance, which is being provided by max-pooling in our approach, is an essential characteristic of the action recognition neural network.
To retain spatial information of joints, we did not merge features along the joint dimension during temporal feature extraction; therefore, no pooling or strided convolutions were used along the spatial dimension.

\subsection{Capsule Neural Network}\label{caps}
In CapsNet~\cite{sabour2017dynamic}, contrary to the scalar output of neurons in traditional CNN, a \emph{set} of neurons are involved in the construction of a neuron output. The output vector from each capsule, called the instantiation vector, encodes a specific type of entity, such as an action or part of it in the input. The magnitude of the instantiation vector describes the likelihood of the existence of an entity in the input, and its orientation in its high dimensional space describes its characteristics. Capsules can be used in layers where each low-level capsule, called a primary capsule, represents the probability of the presence of a specific pattern in the input. Lower-level capsules provide their output to those at higher levels through dynamic routing. This mechanism proposes an iterative workflow for routing lower-level capsules to higher levels. Regarding $u_i$  as $i_{th}$  lower-layer capsule and $v_j$  as $j_{th}$ higher-level capsule, routing between low-level capsules and higher-level ones take place using algorithm~\ref{algorithm} through $r$ iterations.

\begin{algorithm}[ht]
\label{algorithm}
            $u_{i}$: input
            
            $W_{ij}$: weight
            
            $c_{ij}$: coupling coefficient
            
            initialize the log prior matrix $b_{init}$.
            
            ${\widehat u_{j|i}} = {W_{ij}}{u_i}$
            
            ${c_{ij}} \leftarrow \rm{softmax}({b_{ij}})$

			\For{\emph{r} iteration}
			{${\widehat c_{ij}} \leftarrow \rm{softmax}({b_{ij}})$
			
            $c_{ij}^{new} \leftarrow \eta {\widehat c_{ij}} + \left( {1 - \eta } \right)c_{ij}^{old}$
	
			 ${s_i} \leftarrow \sum\nolimits_i {{c_{ij}}{{\widehat u}_{j|i}}} $

            ${v_j} \leftarrow \rm{squash}({s_j})$
            
            ${b_{ij}} \leftarrow {b_{ij}} + {\widehat u_{j|i}}.{v_j}$	}
\caption{Dynamic Routing Algorithm}
\end{algorithm}

In algorithm~\ref{algorithm}, the C matrix describes the contribution strength of each low-level concept to a high-level one and is updated according to the soft updating rule~\cite{lin2018learning} to prevent early over-routing. Since a capsule output orientation indicates characteristics of an entity that is being represented by the capsule, the non-linearity applied to its output should not affect output orientation, but only change output length which indicates the probability of the presence of an entity.
To this end, the squash activation function is applied to the output of each capsule.

\begin{equation}
    \rm{squash}({s_j}) = \frac{{{{\left\| {{s_j}} \right\|}^2}}}{{1 + {{\left\| {{s_j}} \right\|}^2}}}\cdot \frac{{{s_j}}}{{\left\| {{s_j}} \right\|}}
    \label{squash}
\end{equation}

By using the Squashing non-linearity, described in equation~\ref{squash}, short vectors can be reduced to almost zero length, and long vectors can be lengthened to slightly lower than unit length while retaining their orientation. Therefore, the squash activation function is applied to the output of the capsule.

\begin{figure}[!ht]
    \centering
    \includegraphics[width=0.7\columnwidth,scale=1]{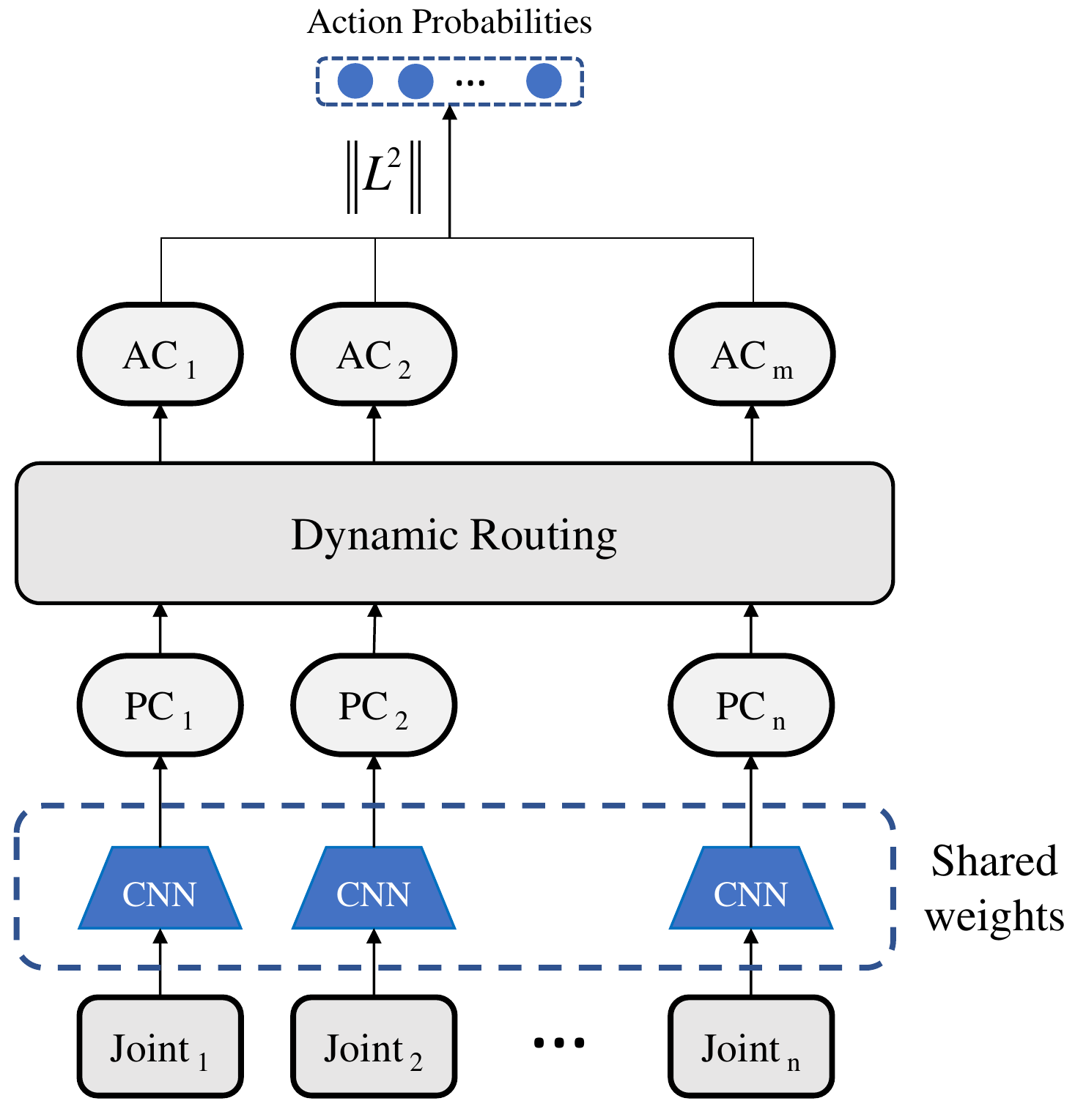}
    \caption{The Overall pipeline of our approach, where PC and AC denote the Primary Capsules and Action Capsules, respectively.}
        \label{fig:pipeline}
\end{figure}

In our approach, we take advantage of CapsNet in extracting high-level relationships between temporal features of joints. To do this, as shown in ~\autoref{fig:pipeline}, we assign a low-level capsule for each joint. Therefore, each low-level capsule will encode temporal features of a specific joint into a vector. Each high-level capsule (called action capsule in this work) represents an action class. Therefore, the length of each action capsule indicates the likelihood of the presence of action in a given sequence. 

The dynamic routing process detailed in algorithm~\ref{algorithm}, which aggregates information from primary capsules and passes it to action capsules, is an important part of our attention mechanism because of the following reasons:
(1) The routing controls the flow of information from primary capsules to action capsules. This is accomplished by weighting the output vector of primary capsules using coupling coefficients. As a result, the network attends to a specific set of joints to decide on the presence of action. 
(2) The routing is dynamic, which means it occurs at inference and alters with each sample. During this iterative procedure joints that are contributing to the activation of wrong actions will be weeded out more through each iteration while the information from useful joints will be emphasized. Although, the model is robust to slight changes in the way actions are executed through different joints by each individual.

\subsection{Network Architecture}\label{act-caps}
\begin{figure}[!ht]
    \centering
    \includegraphics[width=0.95\columnwidth,scale=0.8]{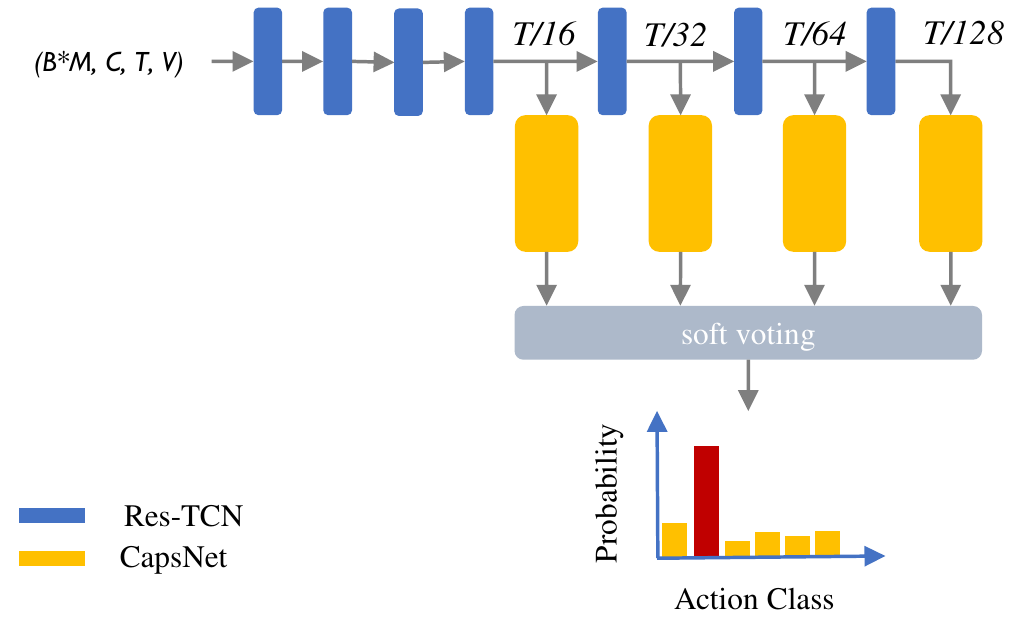}
    \caption{Overall structure of the proposed method. The input skeleton data is reshaped and fed to four consecutive layers of Res-TCN to capture the dynamic features of each joint. At the four late stages, outputs of Res-TCN layers are given to Capsule Networks. To obtain the final prediction, soft voting is applied to predictions of capsules. $B$ denotes batch size and $C, T, V$, and $M$ are the number of channels, frames, joints, and subjects, respectively.}
    \label{overall_architecture}
\end{figure}

In this section, we introduce capsule-based activity recognition to embed relationships between joints in temporally encoded features. ~\autoref{overall_architecture} illustrates the proposed structure. In the first step, we feed the input to a series of Res-TCN layers whose weights are shared across all joints. The late layers of Res-TCNs are passed to a multi-stage activity capsule network, which consists of different activity capsule networks each of which makes predictions based on the temporal features received from a single Res-TCN layer. The reason for using multi-stage architecture is that applying each Res-TCN leads the action capsules to consider larger temporal horizons in the input data. The final prediction of the model is obtained by summing all the predictions of the capsules.

\begin{figure*}[!ht]
    \centering
        \includegraphics[width=1\linewidth]{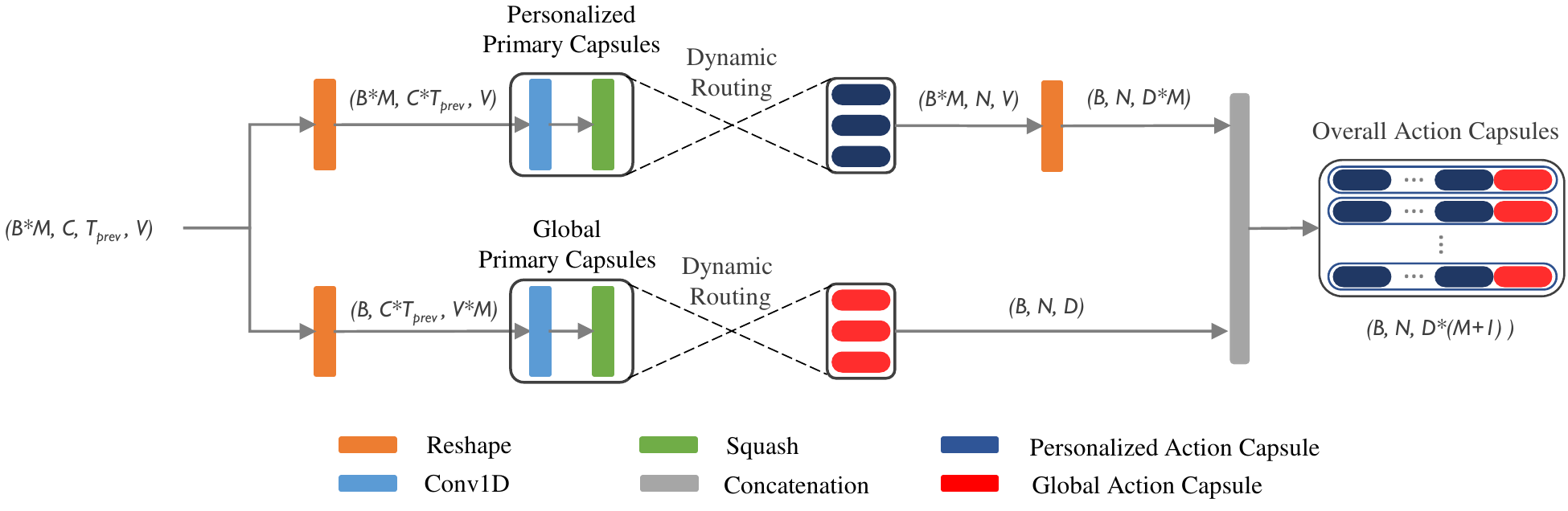}
\caption{Structure of a proposed capsule neural network, where $B$ denotes the batch size, $C, T_{prev}, V$, and $M$ denote the number of channels, frames, joints, and subjects, respectively. $N$ and $D$ represent the number of action classes and the dimension of the instantiation vector of each action capsule, respectively.}
\label{fig:net}
\end{figure*}

Up to this point, we have considered the recognition of actions done by a single subject. However, subjects often perform mutual activities, and data from the skeletons of both subjects participating in the action are needed to recognize the action. As an example, using only skeleton data of a single person makes it difficult to differentiate between activities such as 'kicking' and 'kicking something'. To this end, we define two types of capsule networks: 
(1) \textbf{Personalized capsules}: consider skeleton data from only one subject, thus, As shown in the upper stream of ~\autoref{fig:net}, primary capsules are fed with the features of a single skeleton. Personalized action capsules are constructed by aggregating the personalized primary capsules.
(2) \textbf{Global capsule}: explores the joint relationships between individuals and derives action capsules by all joints present in action. When using personalized capsules, we are only able to obtain the probability of the presence of actions for each individual separately and we cannot fully infer the full picture of the overall action. This is because each set of capsules extracted from each individual does not consider the movements of the other individual(s) present in the input data. In contrast, global capsules are created by feeding all joints present in the scene into an action capsule module to recognize the overall action. Therefore, with our proposed global capsules, shown in the lower stream of ~\autoref{fig:net}, we can find correlations between any set of joints without taking into account who the joints belong.

Overall, as illustrated in ~\autoref{overall_architecture}, we feed data of all subjects present in the scene to a series of Res-TCN layers and generate primary capsules for all joints. Then, at different stages, we construct a personalized capsule for each individual and a global capsule for all subjects. The final activity capsule in each stage is created by concatenating the individual’s personalized capsule and their global capsules in the instantiation parameter dimension. Finally, we aggregate predictions from different stages by soft voting to get the final prediction.

\section{Results and Discussion}

\subsection{Implementation Details}
All experiments are conducted on an RTX 3050 TI GPU with PyTorch deep-learning framework. To train our models, we use Adam optimizer with an initial learning rate of 0.001 and divided by 10 in the 30th epoch and 50th epoch. We use 10 epochs of warm-up at the start of training and stopped the training at the 60th epoch while the batch size was set to 32. To optimize the network hyper-parameters, we use Optuna~\cite{akiba2019optuna}.

For the NTURGB+D dataset, we perform the preprocessing proposed by~\cite{shi2020decoupled}. The maximum number of frames for each sample is 300, and for samples having less than 300 frames, we pad the remaining frames with zero. For input data, 150 frames are sampled uniformly and then centrally cropped to 128 frames. In this dataset, there are at most two subjects in each sample. We pad the second body with zero. Normalizing data is done by taking joint 2 as the origin and calculating the coordinates of the other joints based on it.

\subsection{Datasets}
\textbf{NTURGB+D}~\cite{shahroudy2016ntu} is a large-scale dataset for 3D skeleton-based action recognition. It contains 56880 samples performed by 40 subjects aged between 10 and 35. The dataset contains 60 action classes consisting of 40 daily actions, 9 health-related actions, and 11 mutual actions. The samples contain a sequence of 3D spatial coordinates of 25 joints of the human skeleton, and each sample contains a maximum of two subjects and 300 frames for each subject.
It is recommended by~\cite{shahroudy2016ntu} that two protocols be followed to evaluate proposed methods: (1) Cross Subject (X-sub) evaluation: Data from 20 subjects is used for training, and the remaining data from the other 20 subjects are used for testing. (2) Cross View (X-view) evaluation: data from camera views 2 and 3 are used for training, and data from camera view 1 is used for testing.

\textbf{Northwestern-UCLA} dataset
\cite{wang2014cross}
contains human skeleton data from three different viewpoints, captured simultaneously by Kinect cameras. There are 1494 video clips in 10 categories each performed by 10 different subjects. Our evaluation protocol is the same as in~\cite{wang2014cross}. The samples from the first two cameras serve as training data and those from the third camera serve as testing data.

\subsection{Comparison with the state-of-the-Art Methods}
We compare our model with the state-of-the-art skeleton-based action recognition methods on both the NTURGB+D~\cite{shahroudy2016ntu} dataset and N-UCLA~\cite{wang2014cross} dataset. The results of these two comparisons are shown in ~\autoref{tab:ntu} and ~\autoref{tab:nucla}, respectively. From ~\autoref{tab:ntu}, it can be seen that in terms of accuracy, our method achieves competitive results in both evaluation benchmarks with 96.3\% in Cross-View and 90.0\% in Cross-Subject. Additionally, from ~\autoref{tab:GFLOPs}, in terms of computational complexity, our method is significantly more efficient than state-of-the-art methods with 3.84 GFLOPs which constitutes 61.6\% less required computation than 4s-Shift GCN~\cite{cheng2020skeleton}. 

As shown in ~\autoref{tab:nucla}, the proposed \emph{Action Capsules} outperforms the state-of-the-art methods by achieving an accuracy of 97.3\% in N-UCLA dataset~\cite{wang2014cross}. Additionally, from ~\autoref{tab:GFLOPs}, it has the lowest computational cost of 0.6 GFLOPs compared to existing approaches.

\begin{table}[!ht]
    \centering
    \caption{Comparison with the SOTA Methods: NTURGB+D Dataset.}
    \label{tab:ntu}   
    \fontsize{9}{11}
    \begin{tabular}{c c c}
    \toprule
     \textbf{Methods}    &  \textbf{CV (\%)} & \textbf{CS (\%)}\\
    \midrule
    HBRNN-L~\cite{du2015hierarchical} & 64.0 & 59.1\\
    STA-LSTM~\cite{song2017end} & 81.2 & 73.4\\
    GCA-LSTM~\cite{liu2017global} & 82.8 & 74.4\\
    VA-LSTM~\cite{zhang2012microsoft} & 87.6 & 79.4\\
    \midrule
    HCN~\cite{li2018co} & 91.1 & 86.5\\
    3SCNN~\cite{liang2019three} & 93.7 & 88.6\\
    \midrule
    ST-GCN~\cite{yan2018spatial} & 88.3 & 81.5\\
    AS-GCN~\cite{li2019actional} & 94.2 & 86.8\\
     2s AGC-LSTM~\cite{si2019attention}    &  95 & 89.2\\
     MS-AAGCN~\cite{shi2020skeleton} & 96.2 & 90.0\\
     4s Shift-GCN~\cite{cheng2020skeleton} & 96.5 & 90.7\\
     CTR-GCN~\cite{chen2021channel} & 96.8 & 92.4\\
     \midrule
     \midrule
    \textbf{Action Capsules (ours)} & \textbf{96.3} & \textbf{90.0}\\
    \bottomrule
    \end{tabular}
    
\end{table}
\begin{table}[!ht]
    \centering
    \caption{Comparison of Classification Accuracy with the SOTA Methods: N-UCLA Dataset.}
    \label{tab:nucla}
    \fontsize{9}{11}
    \begin{tabular}{c c}
    \toprule
     \textbf{Methods}  &  \textbf{Top-1 (\%)}\\
     \midrule
     Actionlet Ensemble~\cite{wang2013learning} & 76.0\\
     HBRNN-L~\cite{du2015hierarchical} & 78.5\\
     Ensemble TS-LSTM~\cite{lee2017ensemble} & 89.2\\
    2s-AGC-LSTM~\cite{si2019attention}    &  93.3\\
    4s-Shift-GCN~\cite{cheng2020skeleton}  &  94.6\\
    CTR-GCN\cite{chen2021channel} & 96.5\\
     \midrule
     \midrule
    \textbf{Action Capsules (ours)} &  \textbf{97.3}\\
    \bottomrule
    \end{tabular}

\end{table}

\begin{table}[!ht]
    \centering
    \caption{Comparison of GFLOPs with the SOTA Methods}
    \label{tab:GFLOPs}
    \fontsize{9}{11}
    \begin{tabular}{c c c}
    \toprule
     \textbf{Methods}  &  \textbf{NTU} & \textbf{N-UCLA}\\
     \midrule
    2s-AGC-LSTM~\cite{si2019attention}    &  54.4 & 10.9\\
    4s-Shift-GCN~\cite{cheng2020skeleton}  &  10 & 0.7 \\
    CTR-GCN~\cite{chen2021channel} & 15.4 & 2.52 \\
     \midrule
     \midrule
    \textbf{Action Capsules (ours)} & \textbf{3.48} &  \textbf{0.6}\\
    \bottomrule
    \end{tabular}
    
\end{table}
\section{Visualization and Discussion}\label{visualize}
\subsection{Iterative Mechanism}
As mentioned in ~\autoref{caps}, coupling coefficients of capsules are computed by an iterative mechanism called dynamic routing. The objective of this section is to visualize this process and determine the optimal number of routing iterations.

\subsubsection{Activation of capsules through dynamic routing}
Referring to algorithm~\ref{algorithm}, during inference at the routing stage, the log prior matrix $b$ is initialized with the values learned during training and updated a pre-determined number of times to refine the coupling coefficients. Therefore, key joints for the performed action should get a higher weight of coupling coefficients. 

\begin{figure}[!ht]
    \centering
  \subfloat[iteration 1\label{iter1}]{%
       \includegraphics[scale=1, width=0.9\linewidth]{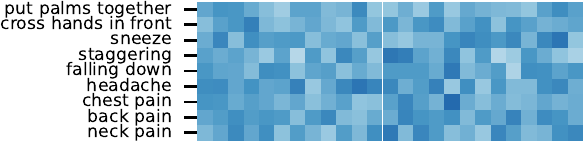}}
       \hfill
       \\
  \subfloat[iteration 2\label{iter2}]{%
        \includegraphics[scale = 0.8,width=0.9\linewidth]{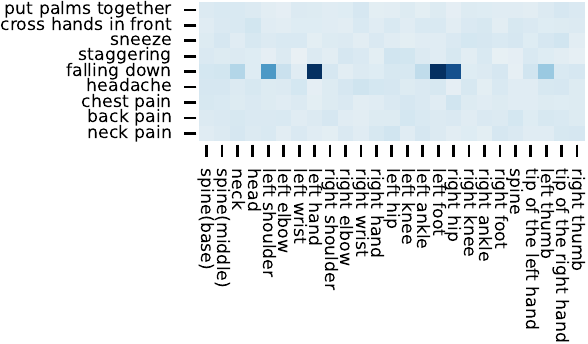}}
    \caption{A cropped illustration of coupling coefficients. Updating of coupling coefficients of capsules when action class 'falling down' is given as input. Rows correspond to high-level capsules (class labels), and columns correspond to low-level capsules (joints).}
    \label{dynamic routing vis}
\end{figure}

~\autoref{dynamic routing vis} shows updating process for the coupling coefficients of stage 4 capsules in each iteration of dynamic routing for a sample of action class "falling down" from NTURGB+D~\cite{shahroudy2016ntu} dataset. In ~\autoref{iter1}, coupling coefficients are nearly the same at the first iteration of routing and another iteration is required to refine them. The second iteration of routing, shown in ~\autoref{iter2}, illustrates that the coupling coefficients related to all classes are adjusted to nearly zero except for the correct action class. In addition, only a few joints are activated in the correct action class. Specifically, joints of ‘left shoulder’, 'left hand', 'left foot', and 'right hip' are related to action class 'falling down' more than other joints. This shows our method can gloss over irrelevant information and attend to a specific set of important joints to classify actions.

\begin{figure}[!ht]
    \centering
  \subfloat[stage 1 action capsules\label{c_s1}]{%
      \includegraphics[scale=0.7,height=3.25cm, width=0.9\linewidth]{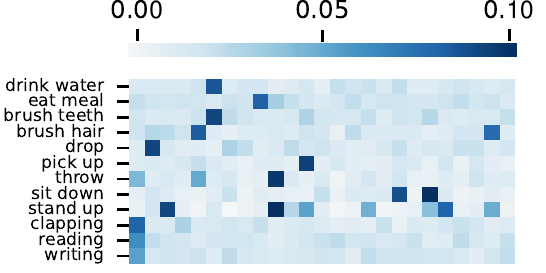}}
      \hfill
  \subfloat[stage 2 action capsules\label{c_s2}]{%
        \includegraphics[scale=0.5,height=4.75cm, width=0.9\linewidth]{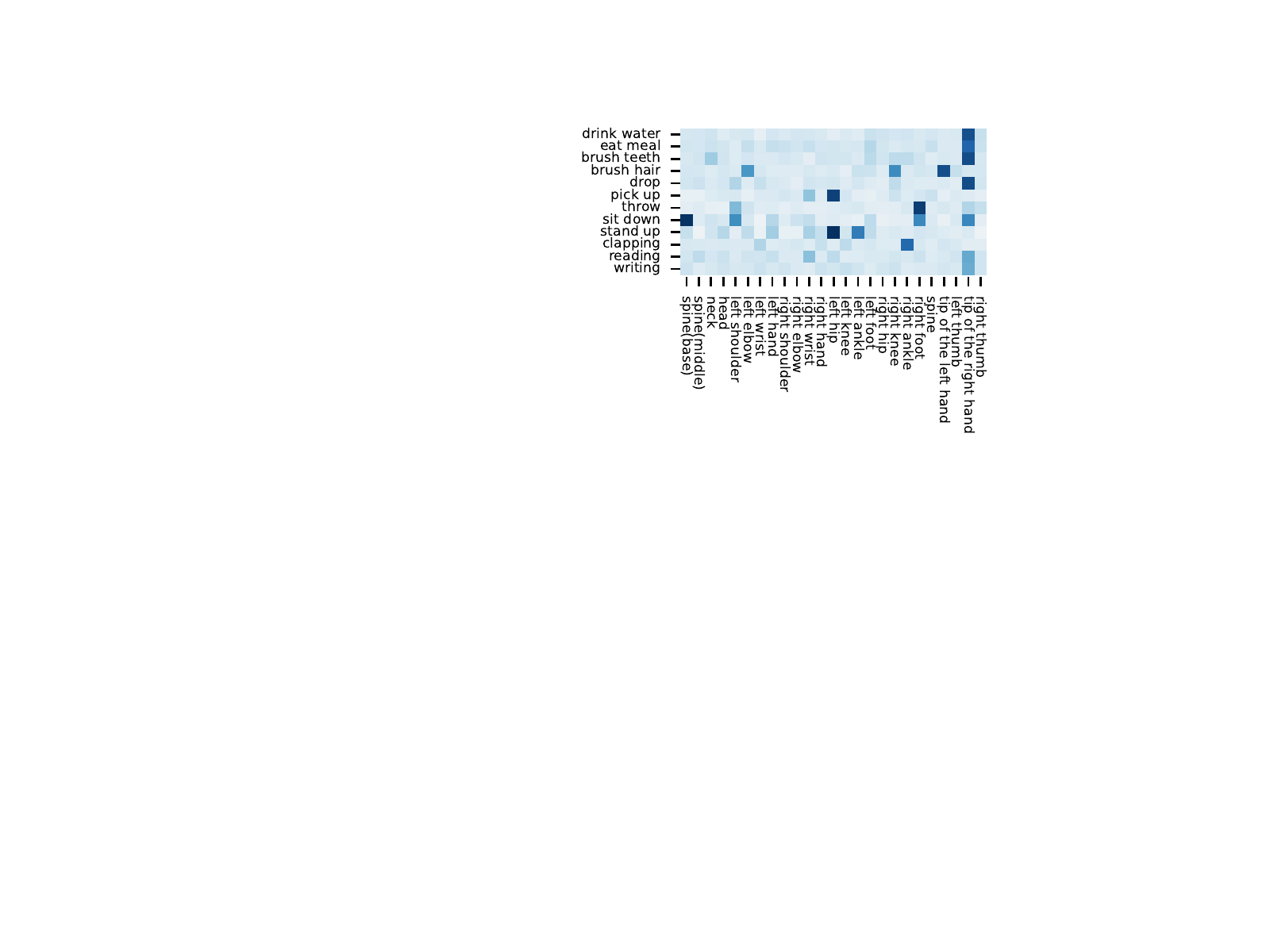}}
        \hfill
        \caption{A cropped illustration of consistency map.}
        \label{consistency}
\end{figure}

\subsubsection{Optimal number of routing iterations}
In this part, we examine the updating of the coupling coefficients when using different numbers of routing iterations to determine whether the refinement process continues with increasing iterations and how it affects the overall performance of the network. As shown in ~\autoref{dynamic routing vis}, dynamic routing alters capsules mostly in the second iteration of the routing, and excessive routing iterations negatively impact the effectiveness of the method and degrade its performance. This may be the result of a high number of routing iterations leading to over-routing. This causes the network to overestimate the contribution of the least prominent joints in higher iterations. 

This is verified in \autoref{tab:4}, which examines the accuracy of the network for different numbers of routing iterations, where two iterations yield higher accuracy for NTURGB+D~\cite{shahroudy2016ntu} dataset.

\begin{table}[!ht]
\caption{The effect of different numbers of routing iterations on our network}
     \centering
     \begin{tabular}{c c}
     \toprule
      \textbf{Routing iterations}  &  \textbf{Accuracy (\%)}\\
     \midrule
     1 & 95.7\\
     2 & 96.3\\
     3 & 96.2\\
     5& 95.9 \\
   \bottomrule
\end{tabular}
     \label{tab:4}   
\end{table}

\subsection{Consistency}
As mentioned in ~\autoref{act-caps}, an action capsule is built by aggregating the joint features from the primary capsules, We define consistency of the model as the ability of a capsule to attend to a consistent set of joints, by assigning higher coupling coefficients in the routing procedure across all samples of an action. To demonstrate the consistency of the proposed model, we feed all data associated with a particular action as input to the network, and in the resulting coupling coefficient matrix, we select only the vector associated with that action. Then, we create a consistency vector for the action by averaging the resulting vectors over the corresponding samples.
By using the same approach for all actions and concatenating the aforementioned vectors, we visualize the consistency of our capsule network as a whole for the NTURGB+D~\cite{shahroudy2016ntu} dataset in ~\autoref{consistency}.

The consistency map of stage 1 and 2 capsules are shown in ~\autoref{c_s1} and ~\autoref{c_s2}, respectively. These figures show which joints are activated primarily over all samples in each action. For instance, ~\autoref{c_s1} illustrates for the samples of 'pick up', the model focuses more on the right hand. Therefore, the consistency of the model is evident in the utilization of specific non-random joints for recognizing each action.

\subsection{Multi-Stage Capsules}
In our proposed method shown in ~\autoref{overall_architecture}, by applying each Res-TCN layer, the output feature considers a broader temporal horizon in input. By providing the output features of Res-TCN layers to capsules in stages, different capsules will consider the input from different temporal horizons. For instance, stage 1 and stage 4 capsules consider the temporal features of joints within $T/16$ and $T/128$ temporal windows, respectively.

From ~\autoref{c_s1} and ~\autoref{c_s2}, it can be seen that each stage pays attention to different joints for the same action in input.
For instance, ~\autoref{c_s2} illustrates the network focusing largely on the 'tip of the right hand' joint for the action classes 'drink water', 'eat meal', and 'brush teeth', while in ~\autoref{c_s1}, different joints are attended by stage 3 capsules for these action classes.
Furthermore, it shows that when using a single stage of capsules, it is difficult for the network to find different joint sets. However, when using multiple stages, it is possible for the network to attend different sets of joints in other stages. Additionally, the results of ~\autoref{tab:multi-stage} confirm the intuitive observations of ~\autoref{consistency} where the single-stage Action Capsule network shows a diminished performance compared to multi-stage models while the accuracy of the model improves by adding each stage.

\begin{table}[ht]
     \centering
     \caption{The effect of multiple stages of capsules on our network}
     \label{tab:multi-stage} 
     \begin{tabular}{c c}
     \toprule
      \textbf{Model}  &  \textbf{Accuracy (\%)}\\
     \midrule
     Action Capsule (1-stage) & 95.3\\
     Action Capsule (2-stages) & 95.6\\
     Action Capsule (3-stages) & 95.8\\
     Action Capsule (4-stages) & 96.3 \\
   \bottomrule
    \end{tabular}  
\end{table}

\subsection{Robustness in the classification of similar classes}
\begin{figure}[!ht]
    \centering
    \subfloat[put on a hat\label{put on hat}]{%
       \includegraphics[scale=0.8, width=\linewidth]{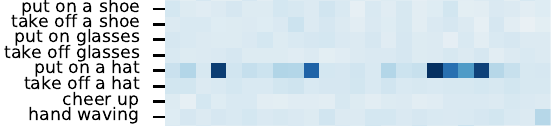}}
       \hfill
       \\
    \subfloat[take off a hat\label{take off hat}]{%
    \includegraphics[scale = 0.8,width=\linewidth]{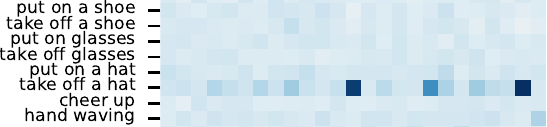}}
    \caption{Coupling coefficients of capsules. The network focuses on a different set of joints in similar classes. Furthermore, joints from classes other than ground truth are not activated.}
    \label{robust similar}
\end{figure}

The skeleton data for some actions are very similar to each other. As an example, in ~\autoref{robust similar}, we compare coupling coefficients of two samples belonging to action classes 'put on a hat' and 'take off a hat', which is challenging to classify using only skeleton data. Despite the similarity of the two samples, our network focuses on a different set of joints in each action class. Moreover, no joints from other classes got activated in both case. Therefore, it is evident that our model learns to attend to different sets of joints for similar and dissimilar actions.

\section{Conclusion}
In this paper, we propose a new Action Capsule Network for skeleton-based action recognition, which to the best of our knowledge is the first attempt at capsule neural networks for this task. It adopts two modules Res-TCN and CapsNet to extract spatial-temporal features. The Res-TCN module is used to learn the temporal dynamics of each joint, followed by a CapsNet which extracts high-level spatial relationships of joints. 
It is confirmed that proposed action capsules can dynamically attend a discriminating set of joints in input during inference, which is unique to each action. Furthermore, we qualitatively and quantitatively demonstrate that multiple stages of capsules improve classification of similar actions to some extent. The final model is evaluated on NTURGB+D and N-UCLA datasets. In NTURGB+D dataset, our network is more efficient than state-of-the-art methods, while achieving comparable accuracy.
In N-UCLA dataset, our proposed method outperforms state-of-the-art methods in terms of accuracy and computational complexity.
\bibliographystyle{unsrt}
\bibliography{biblio.bib}

\begin{thebibliography}{10}

\bibitem{simonyan2014two}
Karen Simonyan and Andrew Zisserman.
\newblock Two-stream convolutional networks for action recognition in videos.
\newblock {\em Advances in neural information processing systems}, 27, 2014.

\bibitem{gaur2011string}
Utkarsh Gaur, Yingying Zhu, Bi~Song, and A~Roy-Chowdhury.
\newblock A “string of feature graphs” model for recognition of complex
  activities in natural videos.
\newblock In {\em 2011 International conference on computer vision}, pages
  2595--2602. IEEE, 2011.

\bibitem{gui2018teaching}
Liang-Yan Gui, Kevin Zhang, Yu-Xiong Wang, Xiaodan Liang, Jos{\'e}~MF Moura,
  and Manuela Veloso.
\newblock Teaching robots to predict human motion.
\newblock In {\em 2018 IEEE/RSJ International Conference on Intelligent Robots
  and Systems (IROS)}, pages 562--567. IEEE, 2018.

\bibitem{bagautdinov2017social}
Timur Bagautdinov, Alexandre Alahi, Fran{\c{c}}ois Fleuret, Pascal Fua, and
  Silvio Savarese.
\newblock Social scene understanding: End-to-end multi-person action
  localization and collective activity recognition.
\newblock In {\em Proceedings of the IEEE conference on computer vision and
  pattern recognition}, pages 4315--4324, 2017.

\bibitem{feichtenhofer2019slowfast}
Christoph Feichtenhofer, Haoqi Fan, Jitendra Malik, and Kaiming He.
\newblock Slowfast networks for video recognition.
\newblock In {\em Proceedings of the IEEE/CVF international conference on
  computer vision}, pages 6202--6211, 2019.

\bibitem{damirchi2020arc}
Hamed Damirchi, Rooholla Khorrambakht, and Hamid~D Taghirad.
\newblock Arc-net: Activity recognition through capsules.
\newblock In {\em 2020 19th IEEE International Conference on Machine Learning
  and Applications (ICMLA)}, pages 1382--1388. IEEE, 2020.

\bibitem{xu2017lie}
Chi Xu, Lakshmi~Narasimhan Govindarajan, Yu~Zhang, and Li~Cheng.
\newblock Lie-x: Depth image based articulated object pose estimation,
  tracking, and action recognition on lie groups.
\newblock {\em International Journal of Computer Vision}, 123(3):454--478,
  2017.

\bibitem{wang2016temporal}
Limin Wang, Yuanjun Xiong, Zhe Wang, Yu~Qiao, Dahua Lin, Xiaoou Tang, and
  Luc~Van Gool.
\newblock Temporal segment networks: Towards good practices for deep action
  recognition.
\newblock In {\em European conference on computer vision}, pages 20--36.
  Springer, 2016.

\bibitem{shahroudy2016ntu}
Amir Shahroudy, Jun Liu, Tian-Tsong Ng, and Gang Wang.
\newblock Ntu rgb+d: A large scale dataset for 3d human activity analysis.
\newblock In {\em Proceedings of the IEEE conference on computer vision and
  pattern recognition}, pages 1010--1019, 2016.

\bibitem{li2018co}
Chao Li, Qiaoyong Zhong, Di~Xie, and Shiliang Pu.
\newblock Co-occurrence feature learning from skeleton data for action
  recognition and detection with hierarchical aggregation.
\newblock {\em arXiv preprint arXiv:1804.06055}, 2018.

\bibitem{yan2018spatial}
Sijie Yan, Yuanjun Xiong, and Dahua Lin.
\newblock Spatial temporal graph convolutional networks for skeleton-based
  action recognition.
\newblock In {\em Thirty-second AAAI conference on artificial intelligence},
  2018.

\bibitem{plizzari2021skeleton}
Chiara Plizzari, Marco Cannici, and Matteo Matteucci.
\newblock Skeleton-based action recognition via spatial and temporal
  transformer networks.
\newblock {\em Computer Vision and Image Understanding}, 208:103219, 2021.

\bibitem{shi2020decoupled}
Lei Shi, Yifan Zhang, Jian Cheng, and Hanqing Lu.
\newblock Decoupled spatial-temporal attention network for skeleton-based
  action-gesture recognition.
\newblock In {\em Proceedings of the Asian Conference on Computer Vision},
  2020.

\bibitem{lea2017temporal}
Colin Lea, Michael~D Flynn, Rene Vidal, Austin Reiter, and Gregory~D Hager.
\newblock Temporal convolutional networks for action segmentation and
  detection.
\newblock In {\em proceedings of the IEEE Conference on Computer Vision and
  Pattern Recognition}, pages 156--165, 2017.

\bibitem{wang2018non}
Xiaolong Wang, Ross Girshick, Abhinav Gupta, and Kaiming He.
\newblock Non-local neural networks.
\newblock In {\em Proceedings of the IEEE conference on computer vision and
  pattern recognition}, pages 7794--7803, 2018.

\bibitem{sabour2017dynamic}
Sara Sabour, Nicholas Frosst, and Geoffrey~E Hinton.
\newblock Dynamic routing between capsules.
\newblock {\em Advances in neural information processing systems}, 30, 2017.

\bibitem{vemulapalli2014human}
Raviteja Vemulapalli, Felipe Arrate, and Rama Chellappa.
\newblock Human action recognition by representing 3d skeletons as points in a
  lie group.
\newblock In {\em Proceedings of the IEEE conference on computer vision and
  pattern recognition}, pages 588--595, 2014.

\bibitem{du2015hierarchical}
Yong Du, Wei Wang, and Liang Wang.
\newblock Hierarchical recurrent neural network for skeleton based action
  recognition.
\newblock In {\em Proceedings of the IEEE conference on computer vision and
  pattern recognition}, pages 1110--1118, 2015.

\bibitem{zhang2017view}
Pengfei Zhang, Cuiling Lan, Junliang Xing, Wenjun Zeng, Jianru Xue, and Nanning
  Zheng.
\newblock View adaptive recurrent neural networks for high performance human
  action recognition from skeleton data.
\newblock In {\em Proceedings of the IEEE international conference on computer
  vision}, pages 2117--2126, 2017.

\bibitem{song2017end}
Sijie Song, Cuiling Lan, Junliang Xing, Wenjun Zeng, and Jiaying Liu.
\newblock An end-to-end spatio-temporal attention model for human action
  recognition from skeleton data.
\newblock In {\em Proceedings of the AAAI conference on artificial
  intelligence}, volume~31, 2017.

\bibitem{liu2017global}
Jun Liu, Gang Wang, Ping Hu, Ling-Yu Duan, and Alex~C Kot.
\newblock Global context-aware attention lstm networks for 3d action
  recognition.
\newblock In {\em Proceedings of the IEEE conference on computer vision and
  pattern recognition}, pages 1647--1656, 2017.

\bibitem{li2017skeleton}
Bo~Li, Yuchao Dai, Xuelian Cheng, Huahui Chen, Yi~Lin, and Mingyi He.
\newblock Skeleton based action recognition using translation-scale invariant
  image mapping and multi-scale deep cnn.
\newblock In {\em 2017 IEEE International Conference on Multimedia \& Expo
  Workshops (ICMEW)}, pages 601--604. IEEE, 2017.

\bibitem{choutas2018potion}
Vasileios Choutas, Philippe Weinzaepfel, J{\'e}r{\^o}me Revaud, and Cordelia
  Schmid.
\newblock Potion: Pose motion representation for action recognition.
\newblock In {\em Proceedings of the IEEE conference on computer vision and
  pattern recognition}, pages 7024--7033, 2018.

\bibitem{ludl2019simple}
Dennis Ludl, Thomas Gulde, and Crist{\'o}bal Curio.
\newblock Simple yet efficient real-time pose-based action recognition.
\newblock In {\em 2019 IEEE Intelligent Transportation Systems Conference
  (ITSC)}, pages 581--588. IEEE, 2019.

\bibitem{shi2020skeleton}
Lei Shi, Yifan Zhang, Jian Cheng, and Hanqing Lu.
\newblock Skeleton-based action recognition with multi-stream adaptive graph
  convolutional networks.
\newblock {\em IEEE Transactions on Image Processing}, 29:9532--9545, 2020.

\bibitem{li2019actional}
Maosen Li, Siheng Chen, Xu~Chen, Ya~Zhang, Yanfeng Wang, and Qi~Tian.
\newblock Actional-structural graph convolutional networks for skeleton-based
  action recognition.
\newblock In {\em Proceedings of the IEEE/CVF conference on computer vision and
  pattern recognition}, pages 3595--3603, 2019.

\bibitem{lin2018learning}
Ancheng Lin, Jun Li, and Zhenyuan Ma.
\newblock On learning and learned representation with dynamic routing in
  capsule networks.
\newblock {\em arXiv preprint arXiv:1810.04041}, 2(7), 2018.

\bibitem{akiba2019optuna}
Takuya Akiba, Shotaro Sano, Toshihiko Yanase, Takeru Ohta, and Masanori Koyama.
\newblock Optuna: A next-generation hyperparameter optimization framework.
\newblock In {\em Proceedings of the 25th ACM SIGKDD international conference
  on knowledge discovery \& data mining}, pages 2623--2631, 2019.

\bibitem{wang2014cross}
Jiang Wang, Xiaohan Nie, Yin Xia, Ying Wu, and Song-Chun Zhu.
\newblock Cross-view action modeling, learning and recognition.
\newblock In {\em Proceedings of the IEEE conference on computer vision and
  pattern recognition}, pages 2649--2656, 2014.

\bibitem{cheng2020skeleton}
Ke~Cheng, Yifan Zhang, Xiangyu He, Weihan Chen, Jian Cheng, and Hanqing Lu.
\newblock Skeleton-based action recognition with shift graph convolutional
  network.
\newblock In {\em Proceedings of the IEEE/CVF Conference on Computer Vision and
  Pattern Recognition}, pages 183--192, 2020.

\bibitem{zhang2012microsoft}
Zhengyou Zhang.
\newblock Microsoft kinect sensor and its effect.
\newblock {\em IEEE multimedia}, 19(2):4--10, 2012.

\bibitem{liang2019three}
Duohan Liang, Guoliang Fan, Guangfeng Lin, Wanjun Chen, Xiaorong Pan, and Hong
  Zhu.
\newblock Three-stream convolutional neural network with multi-task and
  ensemble learning for 3d action recognition.
\newblock In {\em Proceedings of the IEEE/CVF conference on computer vision and
  pattern recognition workshops}, pages 0--0, 2019.

\bibitem{si2019attention}
Chenyang Si, Wentao Chen, Wei Wang, Liang Wang, and Tieniu Tan.
\newblock An attention enhanced graph convolutional lstm network for
  skeleton-based action recognition.
\newblock In {\em proceedings of the IEEE/CVF conference on computer vision and
  pattern recognition}, pages 1227--1236, 2019.

\bibitem{chen2021channel}
Yuxin Chen, Ziqi Zhang, Chunfeng Yuan, Bing Li, Ying Deng, and Weiming Hu.
\newblock Channel-wise topology refinement graph convolution for skeleton-based
  action recognition.
\newblock In {\em Proceedings of the IEEE/CVF International Conference on
  Computer Vision}, pages 13359--13368, 2021.

\bibitem{wang2013learning}
Jiang Wang, Zicheng Liu, Ying Wu, and Junsong Yuan.
\newblock Learning actionlet ensemble for 3d human action recognition.
\newblock {\em IEEE transactions on pattern analysis and machine intelligence},
  36(5):914--927, 2013.

\bibitem{lee2017ensemble}
Inwoong Lee, Doyoung Kim, Seoungyoon Kang, and Sanghoon Lee.
\newblock Ensemble deep learning for skeleton-based action recognition using
  temporal sliding lstm networks.
\newblock In {\em Proceedings of the IEEE international conference on computer
  vision}, pages 1012--1020, 2017.

\end{thebibliography}
\section{Appendix A}
In the \autoref{visualize}, we have visualized cropped illustrations associated with updating coupling coefficients, consistency maps (only stages 1 and 2), and coupling coefficients of two similar classes in Figures 6-8. In this section we provide full visualizations thereof in Figures 9-11.

\onecolumn

\begin{figure*}[!ht]
    \centering
  \subfloat[iteration 1\label{a_iter1-c}]{%
      \includegraphics[height=10.3cm,width=0.46\linewidth]{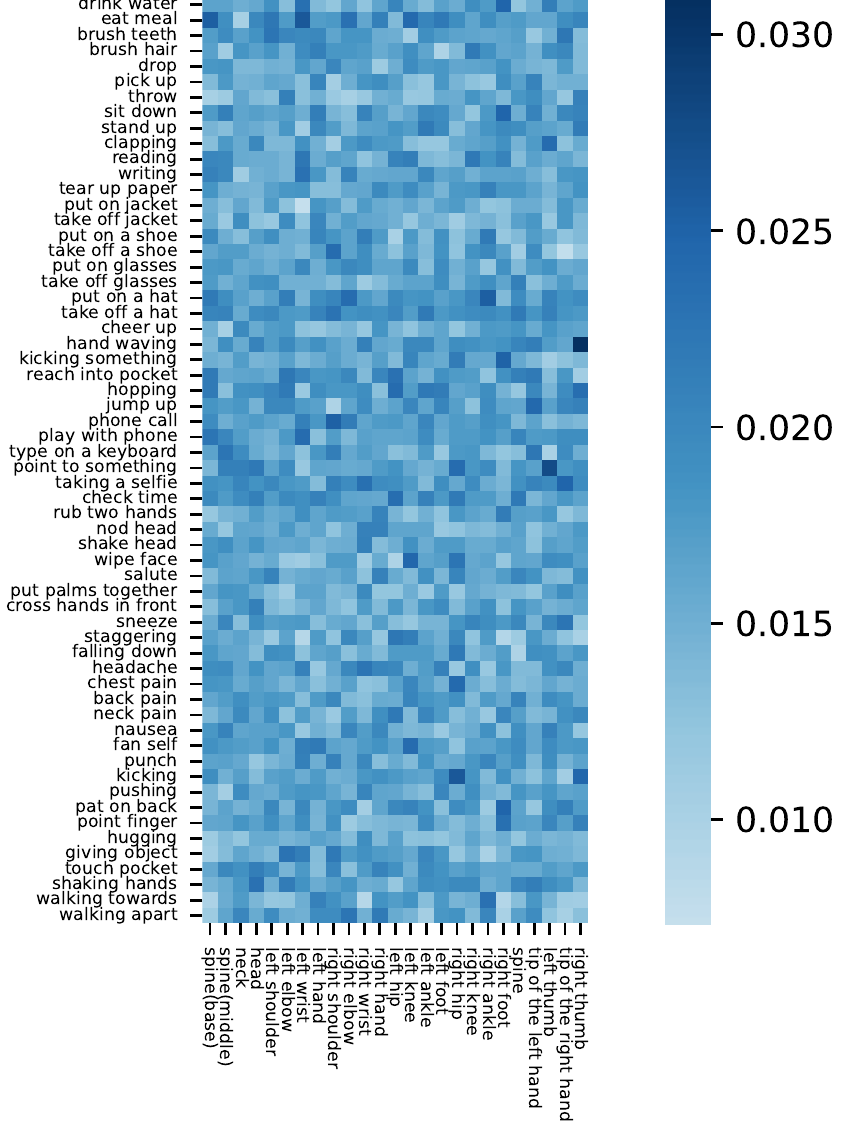}}
      \hfill
  \subfloat[iteration 2\label{a_iter2-c}]{%
        \includegraphics[height=10.3cm,width=0.46\linewidth]{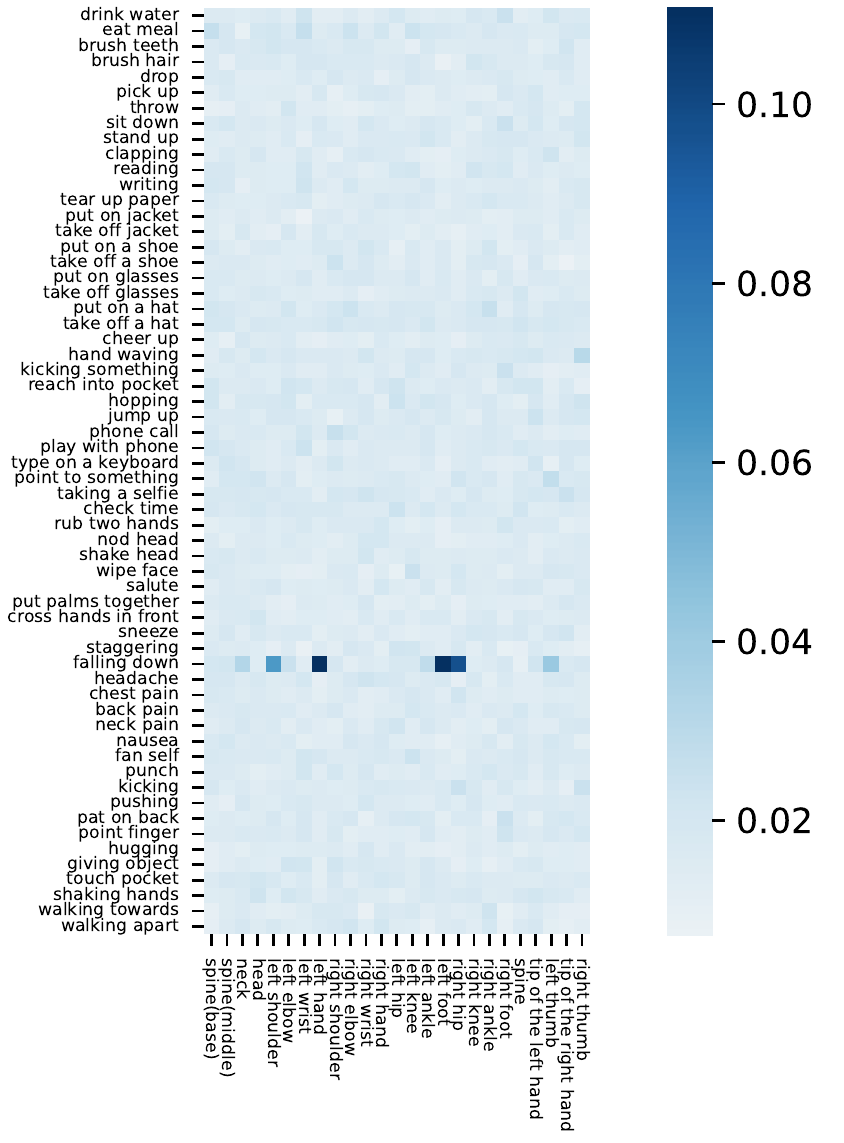}}
        \hfill
        \caption{An illustration of coupling coefficients. Updating of coupling coefficients of capsules when action class ’falling down’ is given as input.}
        \label{a_iteration_c}

        \subfloat[put on hat\label{a_iter1}]{%
      \includegraphics[height=10.3cm,width=0.46\linewidth]{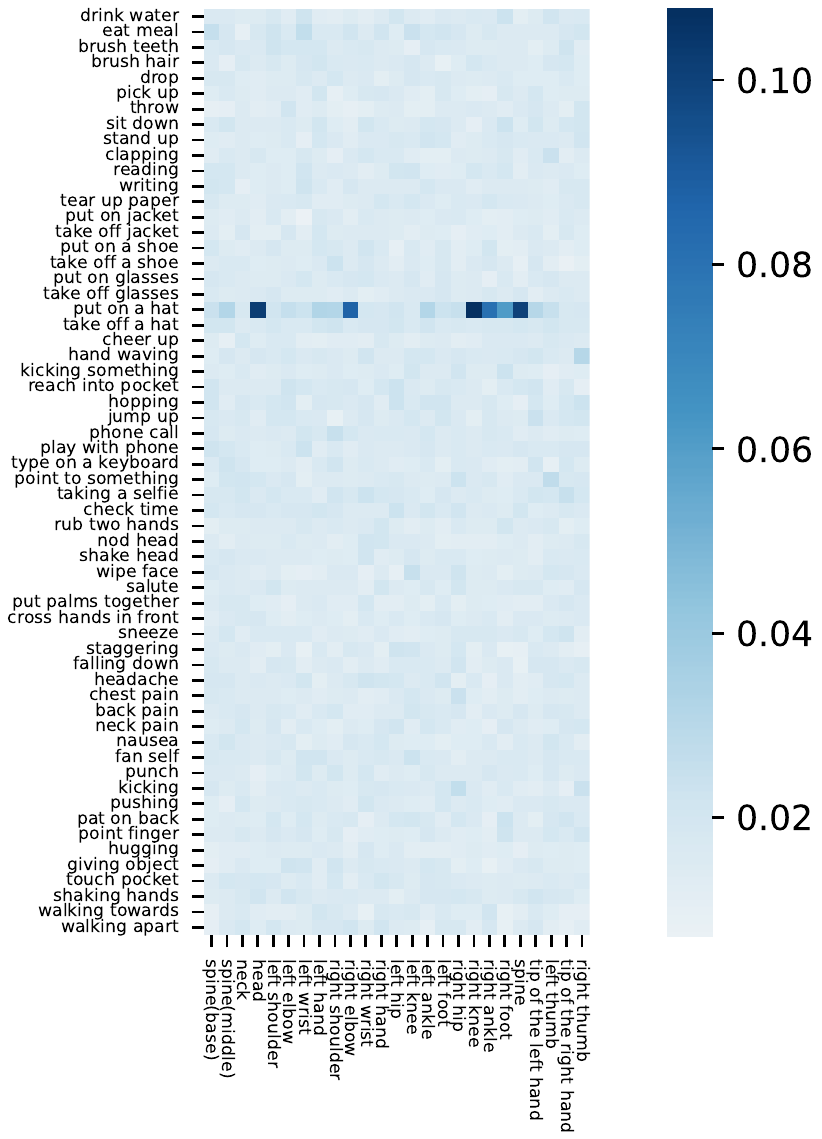}}
      \hfill
  \subfloat[take off hat\label{a_iter2}]{%
        \includegraphics[height=10.3cm,width=0.46\linewidth]{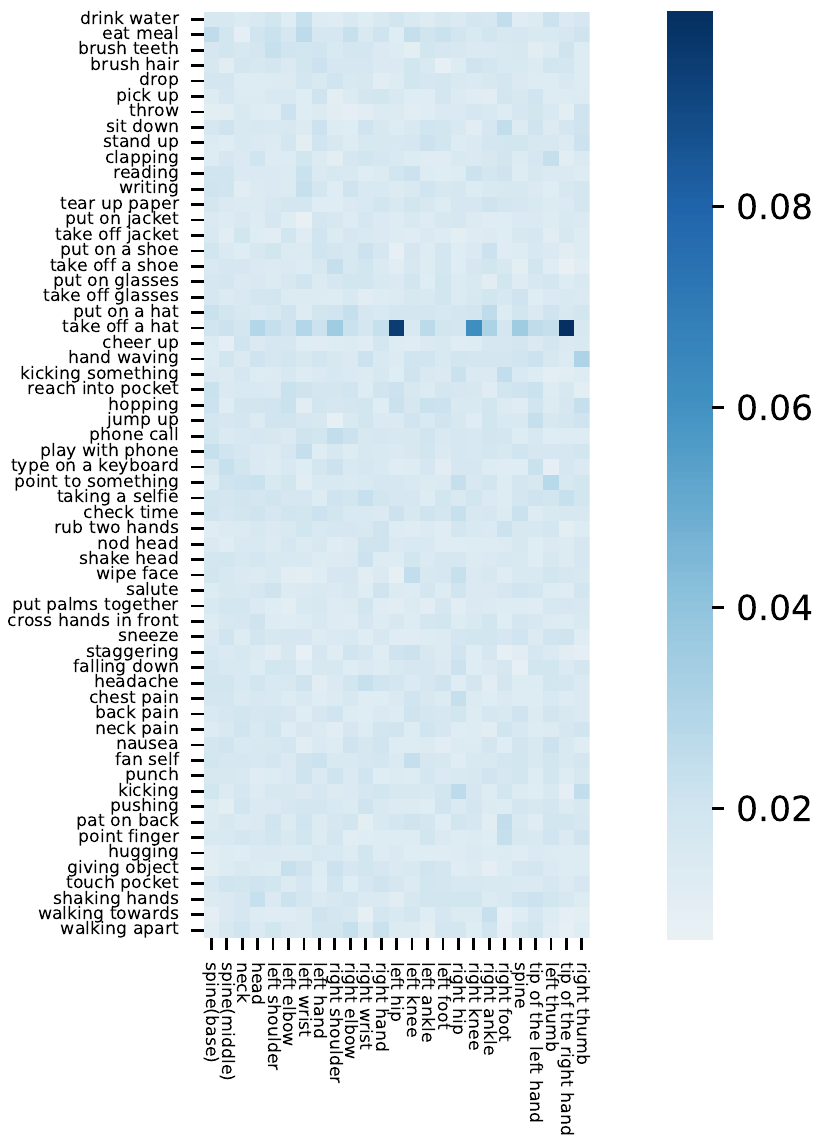}}
        \hfill
        \caption{Comparison of coupling coefficients of capsules in similar classes.}
        \label{a_iteration}
\end{figure*}

\begin{figure*}[!ht]
    \centering
  \subfloat[stage 1 action capsules\label{a_c_s1}]{%
      \includegraphics[height=10.3cm,width=0.46\linewidth]{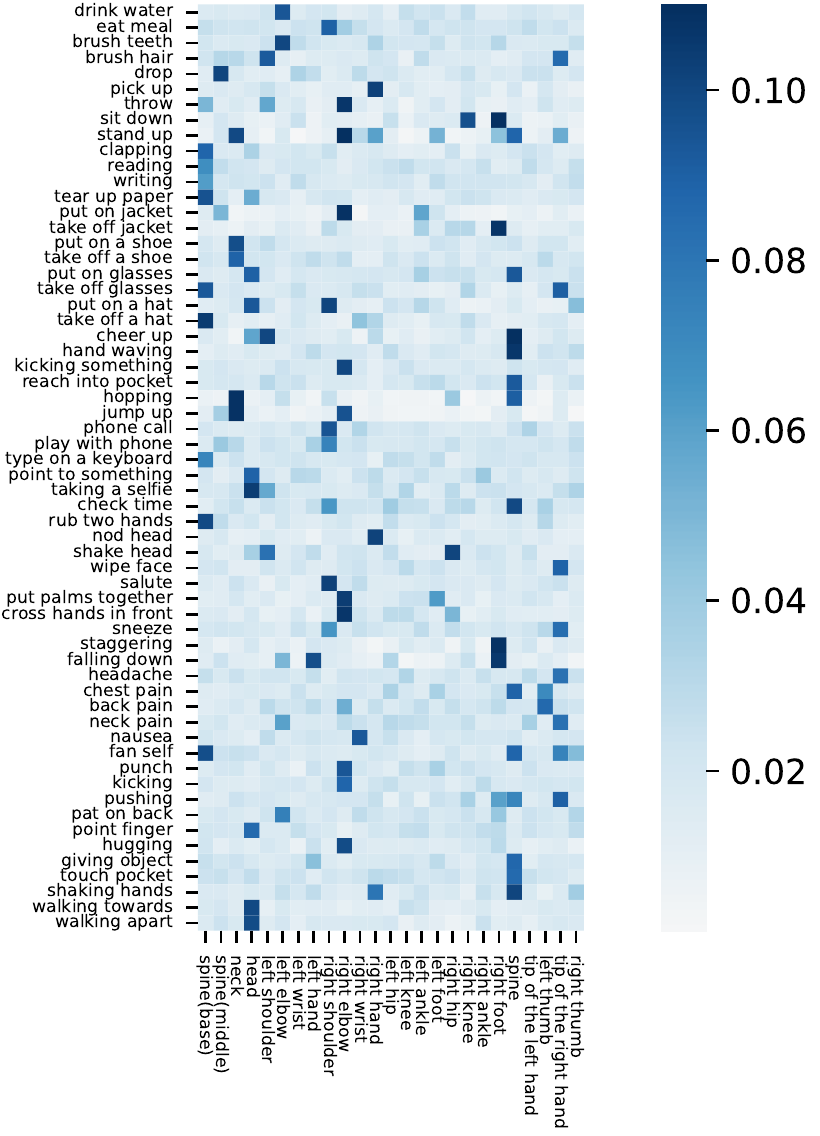}}
      \hfill
  \subfloat[stage 2 action capsules\label{a_c_s2}]{%
        \includegraphics[height=10.3cm,width=0.46\linewidth]{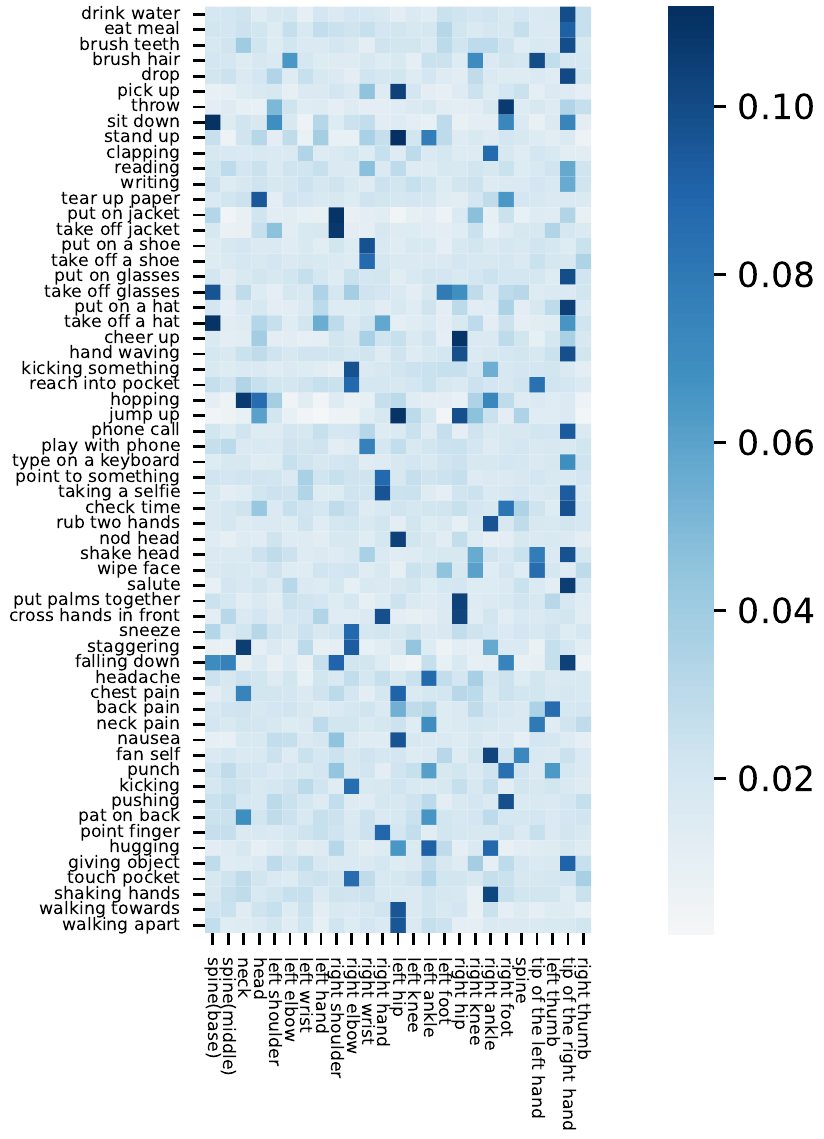}}
        \hfill
        \\
        \subfloat[stage 3 action capsules\label{a_c_s3}]{%
      \includegraphics[height=10.3cm,width=0.46\linewidth]{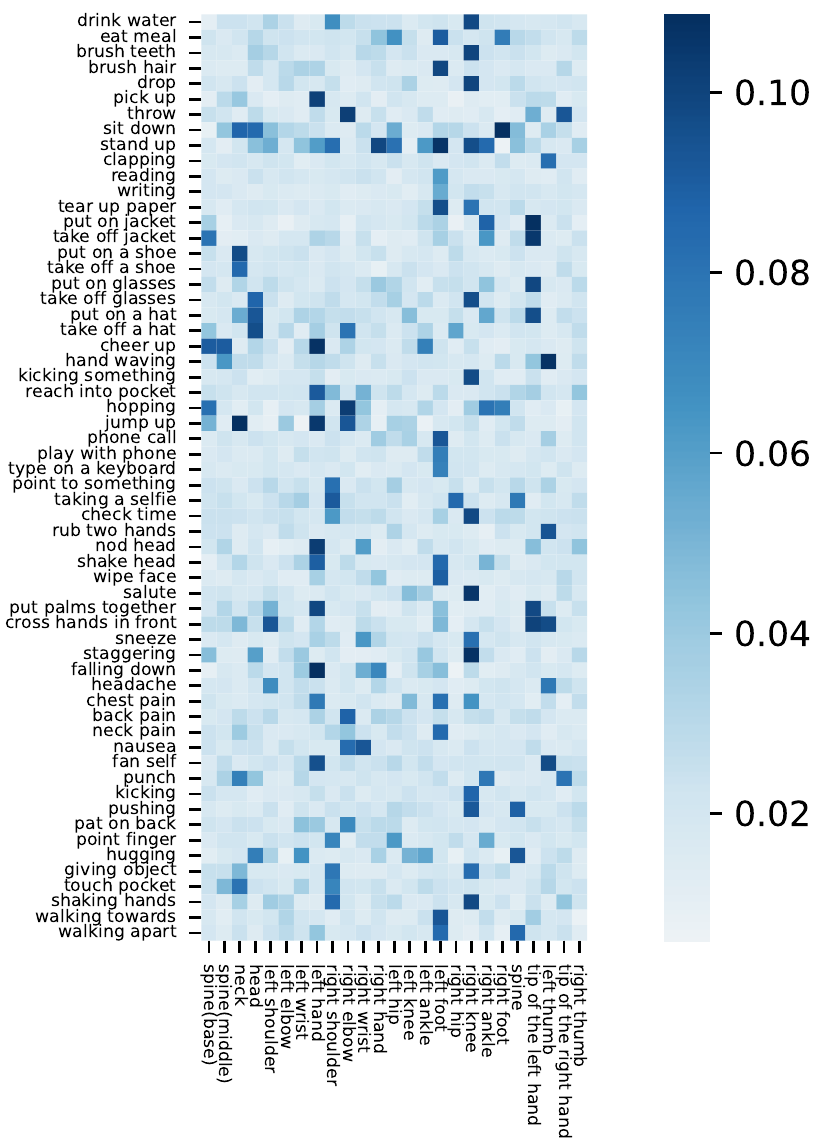}}
      \hfill
  \subfloat[stage 4 action capsules\label{a_c_s4}]{%
        \includegraphics[height=10.3cm,width=0.46\linewidth]{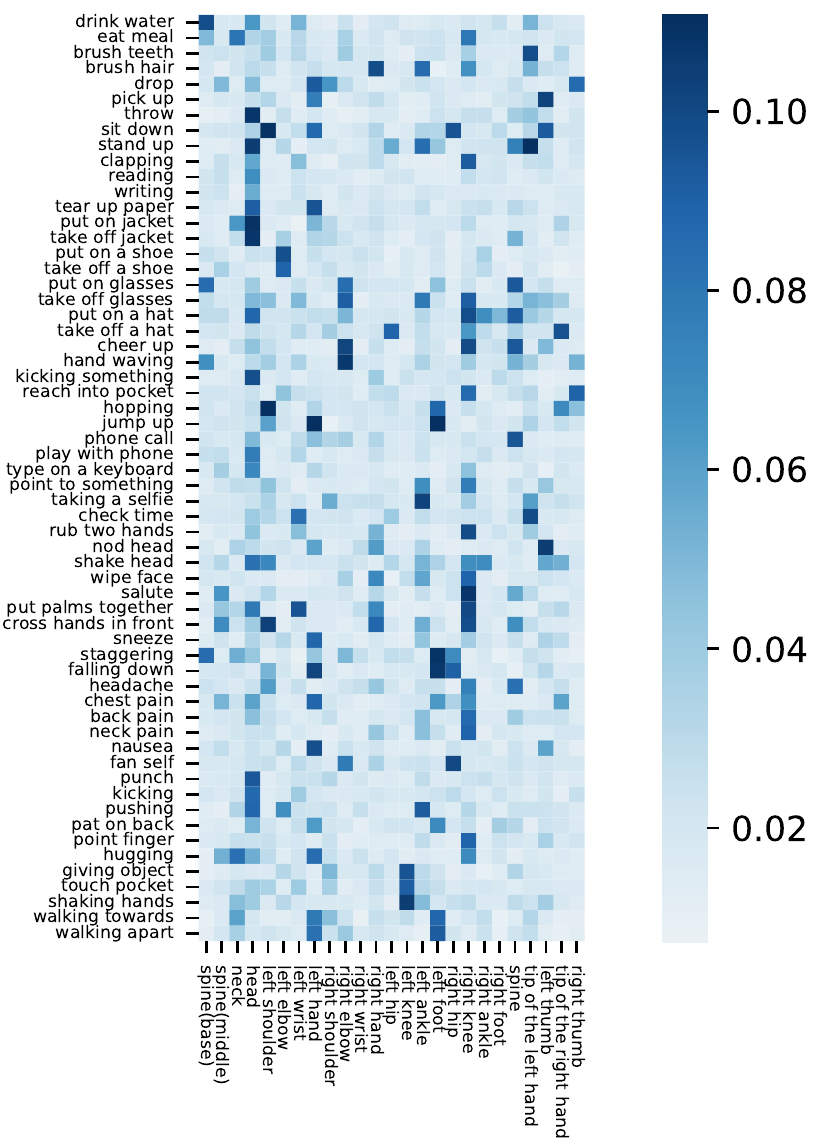}}
        \hfill
        \caption{Consistency map of all stages.}
        \label{a_consistency}
\end{figure*}

\end{document}